\documentclass[runningheads]{llncs}
\usepackage[T1]{fontenc}
\usepackage{graphicx}
\usepackage{xcolor}

\usepackage{csquotes}
\usepackage{amsmath,caption,subcaption,svg}
\usepackage{wrapfig}

\begin{document}
%
\title{Explaining an image classifier with a generative model conditioned by uncertainty}

\titlerunning{Explaining an image classifier with a GAN conditioned by uncertainty}
%


\author{Adrien Le Coz\inst{1,2} \and
Stéphane Herbin\inst{2}\orcidID{0000-0002-3341-3018} \and
Faouzi Adjed\inst{1}\orcidID{0000-0002-0100-9352}}
\authorrunning{A. Le Coz et al.}
%
\institute{IRT SystemX, Palaiseau, France \\
\email{\{adrien.le-coz,faouzi.adjed\}@irt-systemx.fr}
\and
DTIS, ONERA, Université Paris Saclay F-91123 Palaiseau - France
\email{stephane.herbin@onera.fr}}

\maketitle              
\begin{abstract}
Identifying sources of uncertainty in an image classifier is a crucial challenge. Indeed, the decision process of those models is opaque and does not necessarily correspond to what we might expect. To help characterize classifiers, generative models can be used as they allow the control of visual attributes. Here we use a generative adversarial network to generate images corresponding to how a classifier sees the image. More specifically, we consider the classifier maximum softmax probability as an uncertainty estimation and use it as an additional input to condition the generative model. This allows us to generate images that result in uncertain predictions, giving us a global view of which images are harder to classify. We can also increase the uncertainty of a given image and observe the impact of an attribute, providing a more local understanding of the decision process. We perform experiments on the MNIST dataset, augmented with corruptions. We believe that generative models are a helpful tool to explain the behavior and uncertainties of image classifiers.


\keywords{Generative model  \and Explainability of failure conditions \and Uncertainty \and Computer vision.}
\end{abstract}

\section{Introduction}

\textbf{Context: explaining the behavior of image classifiers.}
The growing use of image classifiers in many, sometimes critical, applications (e.g., medical diagnosis, autonomous driving, autonomous aircraft landing) reinforces the need to understand their behaviors. A key issue is to identify the conditions under which such systems are likely to fail, in order to ensure the safety of their use. With this objective in mind, one can consider uncertainty as a measure of potential failure: the question of failure condition identification can be translated into the problem of describing the nature of uncertain data for a given classifier. 

Explainability is currently thought of as a tool to improve the trustworthiness of AI predictive systems. \cite{arrieta2019explainableartificialintelligence,linardatos2021explainableaireview}. In this paper, we propose to provide an explanation of the global classifier behavior as a representation of its uncertain data by using a generative model.

Explainability studies have mainly focused on providing so-called \textquote{post-hoc} explanations that are expected to somehow justify the actual prediction of a trained model. Very few studies have addressed the issue of identifying failure conditions. A related explanatory strategy is the design of counterfactuals \cite{wachter2018counterfactualexplanationsopening,goyal2019counterfactualvisualexplanations}, which aim to identify what minimal and meaningful input modification will lead to a desired prediction change. In particular, several works \cite{zhao2018generatingnaturaladversarial,sauer2021counterfactualgenerativenetworks,lang2021explainingstyletraining,jeanneret2023adversarialcounterfactualvisuala} leverage generative models such as GANs (Generative Adversarial Networks) \cite{goodfellow2014generativeadversarialnetworks} or diffusion models \cite{ho2020denoisingdiffusionprobabilistica}. 
Generative models have also been used to quantify the uncertainty of a classifier \cite{oberdiek2022uqganunifiedmodel} or discover causes of failures \cite{wiles2022discoveringbugsvision,lecoz2022leveraginggenerativemodels}.\\

\noindent\textbf{Main idea: GAN conditioned by the uncertainty of a classifier.} Here we propose to explicitly create a generator of uncertain data. This is done by conditioning a generative model on the uncertainty of a given classifier. Such a generative model can generate infinite amounts of uncertain data (as seen by the classifier) and provides a representation -- an explanation -- of what makes some data hazardous for the classifier. We expect to benefit from the learned model's generalization capacity and use the generative model's latent space -- the \textquote{noise} -- as a compact data representation.

\section{Method}

Generative Adversarial Networks (GANs) \cite{goodfellow2014generativeadversarialnetworks} are a type of generative model known for their generation quality and the controllability offered by their latent (input) space. In particular, they can generate full size images. They are composed of two neural networks: a generator that generates fake images and a discriminator that distinguishes fake images from real ones from the training data. The training is a competition between the two: the generator tries to fool the discriminator, which seeks not to be fooled. During this process, the two improve until the discriminator cannot distinguish any more real from fake data. GAN training is known to be unstable, so an equilibrium has to be found: if the discriminator becomes much better than the generator, it \textquote{wins}, and it's hard for the generator to improve and fill the gap, and vice-versa. Losses and regularizations have been developed to fix the issue \cite{arjovsky2017wasserstein}. After training, we discard the discriminator and use the generator to generate images from noise vectors (\textquote{latent codes}). Interestingly, the model is structured so that interpolations between two latent codes result in a smooth semantic shift of an image into another; for instance, a digit image of \textquote{8} is progressively transformed into a mixture of \textquote{3} and \textquote{8} before ultimately becoming a \textquote{3}, which is not the case if the interpolation is done in the pixel space.

We use the model StyleGAN2 \cite{karras2020traininggenerativeadversarial,karras2020analyzingimprovingimage}, widely used for high-quality face generation and edition. It has a unique architecture. The input latent space $\mathcal{Z}$ is mapped through fully connected layers to an intermediate latent space $\mathcal{W}$. The image is generated progressively at different scales, starting from an initial constant tensor with a size of $4^2$ and $512$ channels, which is up-sampled and transformed by residual convolution layers, and results in images of up to $1024^2$ pixels. Latent codes $\mathbf{w} \in \mathcal{W}$ are transformed into \emph{styles} $\mathbf{s} \in \mathcal{S}$ through learned affine transformations. Those styles will scale the convolution weights of each feature map for each generator layer. Styles applied at low resolution affect high-level aspects of the face (pose, hairstyle...), while at higher resolutions, they affect small details (microstructure...). Latent spaces $\mathcal{W}$ and $\mathcal{S}$ are highly disentangled, meaning that they encode distinct visual attributes along different dimensions. This allows image editing, one attribute at a time \cite{wu2021stylespace}. In particular, we can also use the latent space to characterize classifiers \cite{lang2021explainingstyletraining,oberdiek2022uqganunifiedmodel,lecoz2022leveraginggenerativemodels}. Here, we exploit generative models more straightforwardly by conditioning the generation with the classifier uncertainty, so that it becomes an input of the generator.

\begin{figure}[t]
    \begin{subfigure}[t]{0.63\textwidth}
        \centering
        \includegraphics[width=\linewidth]{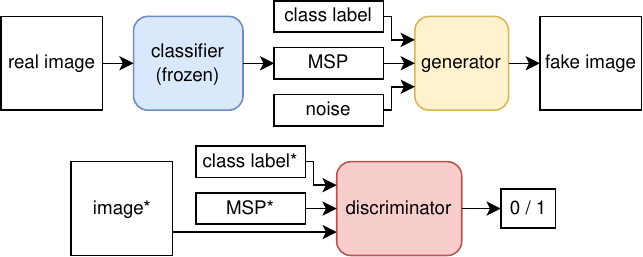}
        \caption{During training time, the additional input MSP conditions the generator. The discriminator evaluates if the combination (class, MSP, image) is realistic. \\
        *for the discriminator, inputs are alternatively (class label, MSP from classifier, real image) and (class condition, MSP condition, fake image generated).} 
        \label{fig:train_gan}
    \end{subfigure}
    \hfill
    \begin{subfigure}[t]{0.33\linewidth}
        \centering
        \includegraphics[width=\linewidth]{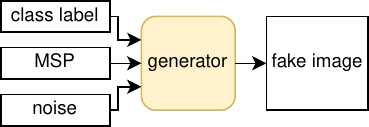}
        \caption{After training, we can generate uncertain images (fix low MSP and vary noise) or identify sources of uncertainty for given images (fix the noise and vary MSP). }
        \label{fig:test_gan}
    \end{subfigure}
    \caption{Training process and structure of the generator.}    
\end{figure}

Our model architecture is depicted in Fig.~\ref{fig:train_gan}.
A conditional GAN \cite{mirza2014conditionalgenerativeadversarial} takes a noise vector as input and a condition. Typically, this can be a one-hot embedding of the class to generate samples of a selected class. A simple way of conditioning a GAN is to concatenate the condition, e.g., encoded as one-hot embedding, to the noise vector as inputs for the generator, and also concatenate the condition to the real or fake image as inputs for the discriminator.

There are several ways to define the prediction uncertainty, e.g., entropy, maximum softmax probability (MSP) \cite{hendrycks2018baselinedetectingmisclassified}, or true class probability \cite{corbiere2021confidenceestimationauxiliary}. We use the imperfect but simple MSP as an uncertainty estimation. We add it as an input condition to the generator. Then after training, the model can generate uncertain data to get a global overview of the uncertainty. We also manipulate data to increase or decrease the uncertainty and exhibit sources of uncertainty. 

MSP values are computed with the classifier (with frozen weights). For the discriminator used on real images, we compute their associated MSP first. For the discriminator used on fake images, we take the MSP used as a condition for the generator, which is not the same as what the classifier would output because of the imperfect generation. 
To condition the generator, we apply the MSPs of random real images using the classifier to track the real distribution of MSP values. Otherwise, the discriminator would more easily make the difference between real and fake data, causing the generator training to be harder.
However, it is important to mention that we do not distinguish between aleatoric and epistemic uncertainty: the generative model is used to sample globally uncertain data.

\section{Preliminary experiments}

\textbf{Two-dimensional \emph{moons} data.} We first illustrate the approach with a simple problem using the moons dataset \cite{pedregosa2011scikitlearnmachinelearning}. The data is 2-dimensional and looks like two interleaving half-circles corresponding to the upper and lower moon classes. The noise level can be adjusted, and we fix it to $0.3$ to have an area where the two classes are mixed. We train a simple fully connected neural network as a classifier. We use a simple generator based on a fully connected network conditioned by one-hot class embedding and the MSP. The network has $5$ layers with a latent space of dimension $8$. The conditioning is a concatenation of the class information as a one-hot embedding vector (of dimension $2$) and the MSP as a continuous value.

\begin{figure}[ht]
    \begin{subfigure}[ht]{0.65\textwidth}
        \centering
        \includegraphics[width=\linewidth]{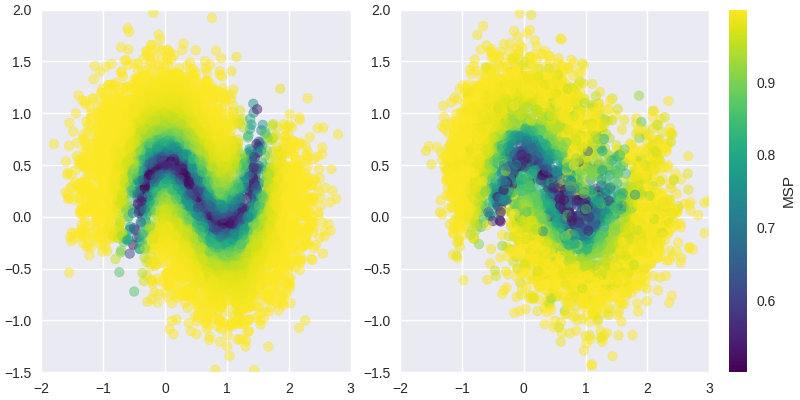}
        \caption{(left) Real data with colors representing the MSP computed by the classifier. (right) Generated data with colors representing the MSP used to condition the generation. The generator captured the meaning of the MSP.}
        \label{fig:results_moons_generator}
    \end{subfigure}
    \hfill
    \begin{subfigure}[ht]{0.32\textwidth}
        \centering
        \includegraphics[width=\linewidth]{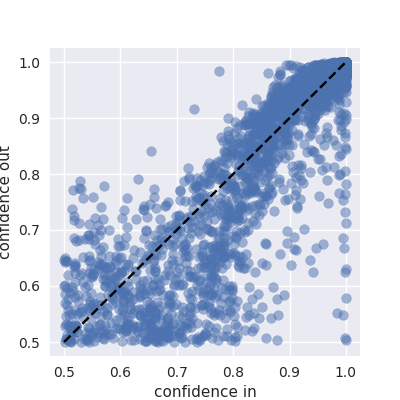}
        \caption{MSP condition (\textquote{in}) vs. MSP computed by classifying the generated data (\textquote{out}).}
      \label{fig:results_moons_generator_MSP}
    \end{subfigure}
    \caption{Qualitative and quantitative results for moons dataset. Uncertainty conditioning works well; the MSP condition corresponds roughly to the real MSP.}

\end{figure}

Fig.~\ref{fig:results_moons_generator} on the left shows the data, with colors representing the MSP obtained when classifying the data. We can see that the MSP is close to $1$, where the classes do not mix but gets lower in the middle area where the classes mix, representing higher uncertainty. We can note that this uncertainty is mostly aleatoric: data can be of either class in the middle region. 
Whereas, in Fig.~\ref{fig:results_moons_generator} on the right shows synthetic data conditioned by MSP. The values are sampled from MSP computed on real data to follow the same distribution. We can see similarities between the locations of real data with high MSP and synthetic data conditioned by high MSP, and likewise for low MSP. The generator captured which kind of data is uncertain and can generate such data when conditioned with low MSP.
\\
For more quantitative results, we follow this process: fix some MSP values as conditions (\textquote{input confidence}), generate fake data, classify it, and obtain the MSP of the classifier (\textquote{output confidence}). Ideally, both values should be the same every time. As seen in Fig.~\ref{fig:results_moons_generator_MSP}, it is not necessarily the case, especially for lower values. Yet, the two are correlated.

\begin{figure}[b!]
    \centering
    \begin{subfigure}[t]{\textwidth}
        \centering
        \includegraphics[width=\textwidth]{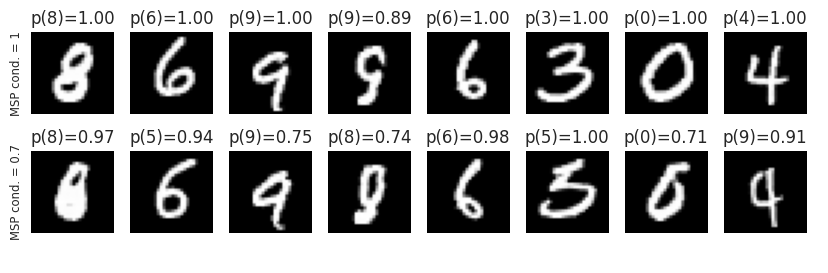}
        \caption{clean MNIST}
        \label{fig:results_mnistClean_generator_new}
    \end{subfigure}
    \vfill
    \begin{subfigure}[t]{\textwidth}
        \centering
        \includegraphics[width=\textwidth]{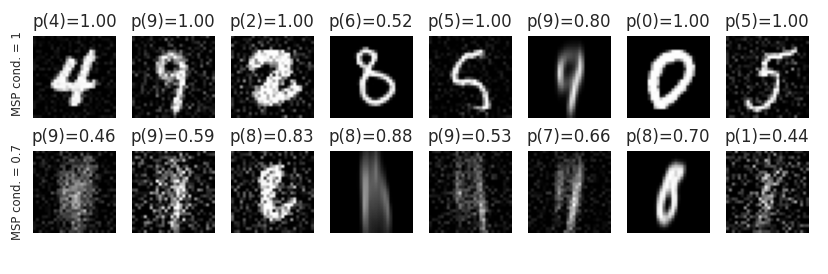}
        \caption{corrupted MNIST}
        \label{fig:results_mnistCorrupt_generator_new}
    \end{subfigure}
    \caption{Samples of images generated with MSP condition fixed at $1$ (top) and $0.7$ (bottom). Above each image is shown the classifier prediction and probability. Images at the bottom look harder, and the classifier is more uncertain.}
\end{figure}

\textbf{MNIST.} Let us now consider more complex data: images. We use the MNIST dataset \cite{deng2012mnistdatabasehandwritten}, which contains black and white images of handwritten digits with ten classes (digits from $0$ to $9$). We train a Convolutional Neural Network to classify digits from images. We consider clean MNIST data, but to make the problem more realistic, we also choose to corrupt MNIST images. We use Gaussian blur and noise similarly to ImageNet-C \cite{hendrycks2019benchmarkingneuralnetwork}. These corruptions are applied on a random half of the images, with a random severity level out of $5$ possible levels.

\begin{wrapfigure}{r}{0.45\textwidth}
\vspace{-0.5cm}
        \centering
        \includegraphics[width=0.45\textwidth]{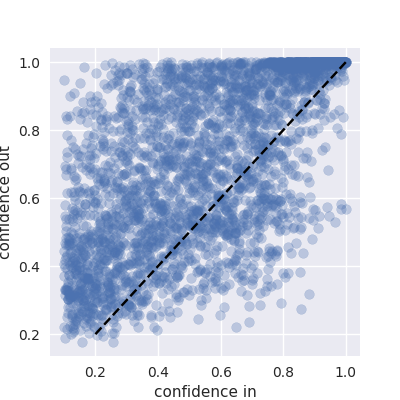}
        \caption{MSP condition (\textquote{in}) vs. MSP computed by classifying the generated data (\textquote{out}).}
        \label{fig:results_mnistCorrupt_generator_MSP}
\vspace{-0.5cm}
\end{wrapfigure}

We found that it results in a reasonable accuracy reduction compared to clean MNIST: now $94.0\%$ on the train set and $93.2\%$ on the validation set (instead of $98.8\%$ and $98.5\%$, respectively). Also, MSP values are more spread out. The GAN is now based on the StyleGAN2 \cite{karras2020analyzingimprovingimage,karras2020traininggenerativeadversarial} architecture to handle images, with additional conditioning for the MSP. We keep the default latent space dimension of $512$ (for the noise), as reducing it makes the training more difficult. 
The conditioning is a concatenation of a class embedding and the MSP value.

We can generate uncertain images by fixing a low MSP value and varying the noise input, as illustrated in Fig. \ref{fig:results_mnistClean_generator_new} and \ref{fig:results_mnistCorrupt_generator_new} bottom. Also, comparing Fig. \ref{fig:results_mnistClean_generator_new} and \ref{fig:results_mnistCorrupt_generator_new} top versus bottom, we gain insight into the classifier's sources of uncertainty by observing what makes given images more uncertain (by fixing noise inputs and lowering the MSP condition). In this case, it is primarily shape, Gaussian noise, and blur that perturbates the classifier. 

The qualitative results in Fig.~\ref{fig:results_mnistClean_generator_new} and~\ref{fig:results_mnistCorrupt_generator_new} show that images generated with the conditioning of MSP $=1$ mainly result in \textquote{output} MSPs close to $1$. We get more spread-out \textquote{output} MSP values when conditioned with MSP $=0.7$.  Fig.~\ref{fig:results_mnistCorrupt_generator_MSP} shows that \textquote{input} MSP and \textquote{output} MSP can be quite different. While not as good as on the moons dataset, we still observe some correlation. We hypothesize that the MSP is much less well-defined on MNIST images than on the moons dataset. More substantial constraints on the conditioning should be considered to improve the results.

\section{Conclusion}
We created an explicit generator of uncertain data that can be used in several ways.
It can give a global outlook of uncertain images by generating them on demand. It can also corrupt images (transform them into their more uncertain form) to visualize sources of local uncertainty. The results are preliminary but encouraging. Leveraging generative models is a promising way to improve explainability when uncertain data is rare. \\
We identified some ideas for future work. The MSP might not contain sufficient information to capture the classifier behavior, so more information, like the full vector, could be considered. The constraint put on the condition during training should be reinforced, for instance with an additional loss term. Furthermore, the MSP might not be calibrated: the probability might be overestimated and require a recalibration.

%
%
%
\newpage

\section*{Acknowledgments}
    This work has been supported by the French government under the "Investissements d'avenir” program, as part of the SystemX Technological Research Institute. \\
    This work was granted access to the HPC/AI resources of IDRIS under the allocation 2022-AD011013372 made by GENCI.

\bibliographystyle{splncs04}
\bibliography{UncertaintyPaper}

\end{document}